\pdfoutput=1

\documentclass[11pt]{article}
\usepackage{changepage,threeparttable} 

\usepackage{tabularx}
\usepackage[final]{acl}

\usepackage{svg}

\usepackage{booktabs}

\usepackage{graphicx} 

\usepackage{times}
\usepackage{latexsym}

\usepackage[T1]{fontenc}

\usepackage[utf8]{inputenc}

\usepackage{microtype}

\title{Location-based Twitter Filtering for the Creation of Low-Resource Language Datasets in Indonesian Local Languages
}

\author{Mukhlis Amien \and Chong Feng \and Heyan Huang \\
        Beijing Institute of Technology \\ Beijing, China \\
\texttt{fengchong@bit.edu.cn} \\}

\begin{document}
\maketitle
\begin{abstract}
Twitter contains an abundance of linguistic data from the real world. We examine Twitter for user-generated content in low-resource languages such as local Indonesian. For NLP to work in Indonesian, it must consider local dialects, geographic context, and regional culture influence Indonesian languages. This paper identifies the problems we faced when constructing a Local Indonesian NLP dataset. Furthermore, we are developing a framework for creating, collecting, and classifying Local Indonesian datasets for NLP. Using twitter's geolocation tool for automatic annotating.

\end{abstract}

\section{Introduction}

Indonesia is the world's fourth most populous country, on December 30, 2021, Indonesia's population is estimated to be 273 million people spread across 17,508 islands~\citep{dukcapil-kemendagri-2022}.  Indonesia is home to around 700 languages, accounting for 10\% of the world's total and ranking second only to Papua New Guinea in terms of language diversity \citep{kohler2019language}. However,  most of these languages are not well mentioned in the literature, many are not taught professionally, and there is no universally accepted standard among speakers of these languages \citep{novitasari-etal-2020-cross}. Due to the growing popularity of "Bahasa Indonesia" as an official language, as the primary written and spoken language throughout the country, a number of these local languages are experiencing a decline in usage. Finally, this pattern may result in the emergence of a nation with a single language \citep{Aji2022}.

\begin{figure}[!htp]
\centering

   \includegraphics[width=0.75\linewidth]{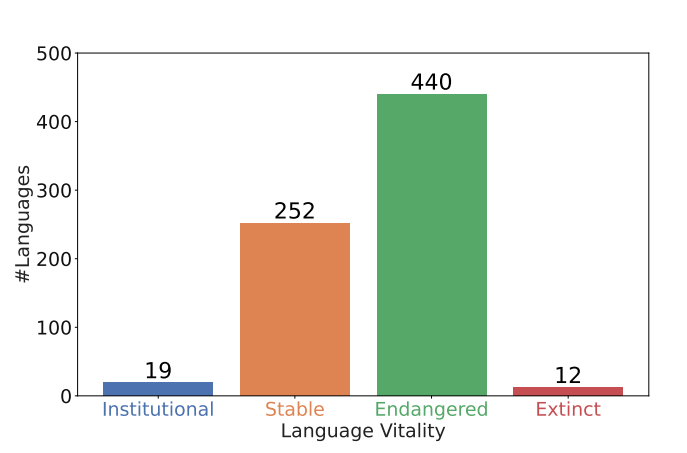}

   \includegraphics[width=0.75\linewidth]{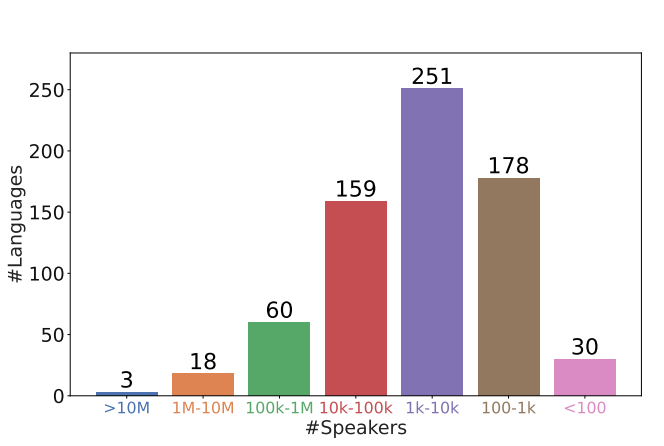}

\caption[Two numerical solutions]{According to Ethnologue, Indonesia has around 700 languages spoken. Up: Language liveliness. Down: Amount of speakers \citep{ethnologue}}
\label{fig:1}
\end{figure}

Many of Indonesia's indigenous languages are endangered, numbering more than 400. According to data from Ethnologue \citep{ethnologue}, which is depicted in figure \ref{fig:1}, there are 440 languages that are categorized as endangered, and 12 that are categorized as extinct. \citet{anindyatri_mufidah_2020} discovered that nearly half of a sample of 98 local Indonesian languages were endangered, and 71 out of 151 local Indonesian languages had fewer than 100k speakers, according to their research.

For the NLP technique to be relevant in the Indonesian environment, it must also consider the dialects of the indigenous languages spoken there. In Indonesia, language dialects are influenced by their speaker's geographical location and regional culture \citep{vander2015dichotomy}. As a result, they can differ significantly in morphology and vocabulary, challenging natural language processing systems. In this study, we will highlight the challenges we encountered when developing a dataset for a natural language processing system utilizing the native Indonesian language. And a framework for creating, collecting, and classifying Indonesian Local Language into datasets for the NLP system. And investigating the geolocation tool on Twitter for automatic annotation. 

Our framework comprises of three phases. The first phase is filtering widely used foreign language tweets from Indonesian local Twitter using fasttext \citep{joulin2016fasttext} to automatically detect language. The second phase is to split the tweet into the formality or informality of an Indonesian text. We choose BERT \citep{https://doi.org/10.48550/arxiv.1810.04805} as our pre-training model since it has been thoroughly researched by Indonesian scholars. The last phase is we try to classify tweets into the area where the tweet was made. According to our findings, getting local language data from the Twitter dataset can be beneficial, but it also presents its own unique set of challenges.


\begin{table}[]
\begin{tabular}{@{}lll@{}}
\toprule
\textbf{Language} & \textbf{ISO} & \textbf{\# Speakers} \\ \midrule
Indonesian        & ind          & 198 M                \\
Javanese          & jav          & 84 M                 \\
Sundanese         & sun          & 34 M                 \\
Madurese          & mad          & 7 M                  \\
Minangkabau       & min          & 6 M                  \\
Buginese          & bug          & 6 M                  \\
Betawi            & bew          & 5 M                  \\
Aceh              & ace          & 4 M                  \\
Banjar            & bjn          & 4 M                  \\
Balinese          & ban          & 3 M                  \\
Palembang (musi)  & mus          & 3 M                  \\ \bottomrule
\end{tabular}
\caption{Displays the number of speakers of Indonesian and the ten most widely spoken local languages in Indonesia acording to Ethnologue \citep{ethnologue} }
\label{tab:1}
\end{table}


\section{Problem Formulation}
The majority of Indonesians speak Indonesian in formal forum, with respected individuals, such as teachers and government officials, as well as with those who do not belong to the group. Because virtually everyone in Indonesia is fluent in the language, and because the Indonesian people consider utilizing Indonesian as a means of communication to be respectable. While the local language is only spoken among the local community, it has numerous variants in written form due to the lack of standardization efforts, making gathering local language data in written form is challenging. As a result, we will design a framework for automatically generating, collecting, and classifying data in 33 Indonesian cities using Twitter geolocation. This is because manually collecting datasets is costly and time-consuming.

\section{Related Works}

\subsection{Multilingual Local Indonesian NLP}

There have been several pretrained multilingual language models proposed recently, including BERT dataset in twitter format \citep{koto2021indobertweet}, mBART \citep{mbartliu}, and mT5\citep{mT5Xue2020}. On the other hand, their coverage is limited to languages that require a lot of resources. In addition to Minangkabau, Sundanese, and Javanese, only mBERT and mT5 include Indonesian native languages, namely Javanese, Sundanese, and Minangkabau, although with just a tiny amount of pretraining data.

The majority of multilingual datasets do not include the Indonesian language, and just a tiny number of them include Indonesian regional languages. One exception is the WikiAnn dataset \citep{pan-etal-2017-cross}, a weakly supervised named entity recognition dataset covering several Indonesian local languages, including Acehnese, Javanese, Minangkabau, and Sundanese. WikiAnn is a weakly supervised named entity recognition dataset created by \citealp{pan-etal-2017-cross}. CommonCrawl; Wikipedia parallel corpora such as MediaWiki Translations and WikiMatix; the Leipzig corpora, which include Indonesian and top five spoken local Indonesian and include  Acehnese, Buginese, Banjar, and Balinese; and JW-300. On the other hand, recent research has raised concerns about the quality of multilingual corpora for minority languages, mainly linguistic diversity. Although a dataset for local languages has been compiled, it does not contain as much information as the ethnologue \citep{ethnologue} claim that there are 700 spoken languages.

\subsection{Local Indonesian Languages NLP System}

Relatively little effort has been done on indigenous languages. Several studies investigated stemming Sundanese \citep{suryani_sunda}; Balinese \citep{subali2019kombinasi} and POS Tagging Sundanese \citep{suryani_sunda}; Balinese \citep{subali2019kombinasi} and Madurese \citep{dewi2020combination}. A comparable corpus of Indonesian Minangkabau words, as well as sentiment analysis tools for Minangkabau, were created by Koto \citep{koto2020towards}. Another group of researchers worked on developing machine translation systems between Indonesian and local languages, such as Dayak Kanayatn \citep{hasbiansyah2016tuning}, Sambas Malay \citep{ningtyas2018penggunaan}, and Buginese \citep{apriani2016pengaruh}. Some researcher investigated the segmentation of Javanese characters in non-Latin scripts. 


\subsection{Local Indonesian Collection Challenge}
\label{cap:challenge}
\subsubsection{Resources are limited}
Insufficient language resources is the most urgent issue with low-resource languages. While there are over 712 languages spoken in Indonesia, the availability of information for each is extremely unequal. For example, Typically, linguistic resources in Indonesian local languages are restricted to formal Indonesian and the top 10 regional languages as in table \ref{tab:1} \citep{ethnologue}. In Indonesia, practically all news organizations and news websites use formal Indonesian, as do communications, legal documents, and books; only a small number of books mention specific local languages. Therefore, the compilation of the dataset itself is an issue and challenging.

\subsubsection{Differences Between Dialects}
Depending on the geographic region, Indonesian native languages can have multiple dialects. Even though they are designated as the same language, the local languages of Indonesian spoken in different regions may differ (have lexical differences) from one another \citep{fauzi2018dialect}. For instance, \citet{anderbeck2012malayic} in table \ref{tab:anderbeck } shown that Jambi Malay dialects vary between villages in the province of Jambi. Similar to Javanese in the Lamongan exhibits up to 13 percent lexical diversity between districts shows in table \ref{tab:javanese} . Similar investigations on other languages, such as Balinese \citep{maharani2018variasi} and Sasak have been undertaken \citep{sarwadi2019lexical}.
\begin{table*}[!htp]\centering
\scriptsize
\begin{tabular}{lrrrrrrrrr}\toprule
English &Suo Suo &Lubuk Telau &Mersam &Teluk Kuali &Mudung Laut &Bunga Tanjung &Pulau Aro &Dusun Teluk \\\midrule
he/she &kau &no &no &no &dio' &no &ino &dio', no \\
I/me &sayo &ambo &awa' &kito, awa' &sayo &ambo &ambo &aku \\
if &bilao &jiko &kalu &kalu &kalu &ko' &kalu &jiko, kalu \\
You &kamu &kamu &kadn &kaan &kau, kamu &ar, kau, kayo &ba'an &kau \\
one &seko' &seko' &seko' &cie' &satu &seko', so &seko' &seko' \\
\bottomrule
\end{tabular}
\caption{The Jambi Malay language has lexical variety among the province's several communities\citep{anderbeck2012malayic}}
\label{tab:anderbeck }
\end{table*}
\begin{table*}[!htp]\centering
\scriptsize
\begin{tabular}{lrrrrr}\toprule
English &Krama &\multicolumn{3}{c}{Ngoko} \\\cmidrule{1-5}
&Eastern &Eastern &Central &Western \\\midrule
Why &punapa &opo'o &ngopo &ngapa \\
Will &badhe &kate, ate &arep &arep \\
I/me &kulo &aku &aku &inyong, enyong \\
You &panjenengan &koen, awakmu, sampeyan &kowe, siro, sampeyan &rika, kowe, ko \\
How &pripun &yo'opo &piye &priwe \\
Not/no &mboten &gak &ora &ora \\
\bottomrule
\end{tabular}
\caption{Javanese dialects and styles vary lexically in different locations of the island of Java. It is requested that native speakers translate the terms}\label{tab:javanese}
\end{table*}

We show further instances of lexical diversity in indigenous Indonesian languages. \citet{maharani2018variasi} and \citet{sarwadi2019lexical} demonstrate lexical differences in Balinese and Sasak, respectively, by requesting translations of frequent words from native speakers. Then, they compared the language of various locales (villages in this case) to one another. The examples are presented in tables \ref{tab:bali_variation} and \ref{tab:sasak }.\citet{pamolango2012geografi} did a similar experiment at 31 observation stations for the Saluan language in the Banggai district of South Sulawesi. While \citet{pamolango2012geografi} did not provide complete instances, they observed a lexical variety of up to 23.5 percent among 200 words from the basic vocabulary.
\begin{table}[!htp]\centering
\scriptsize
\begin{tabular}{lrrrr}\toprule
English &Jimbaran &Unggasan &Kedonganan \\\midrule
Swallow (vb) &Gélék, ngélék &Ngélokang &Gélék, ngélék \\
How &Engken &Kengen &Engken \\
Afternoon &Sanjé &Sanjano &Sanjé \\
Where &Dijé &Di joho &Dijé \\
Scratch (vb) &Gagas &Gauk &Gagas \\
I/me &Tyang &Aku &Tyang \\
Hat &Topong &Cecapil, Tetopong &Capil \\
You &Béné &Éngko &Béné \\
Dawn &Plimunan &Sémongan &Plimunan \\
Cough (vb) &Dékah &Kohkohan &Kokoan \\
Umbrella &Pajéng &Pajong &Pajéng \\
All &Onyé &Konyangan, onyang &Konyangan \\
\bottomrule
\end{tabular}
\caption{Variation in the vocabulary of the Balinese language found in various communities located in the South Kuta district of Bali (Maharani and Candra, 2018)}\label{tab:bali_variation}
\end{table}
\begin{table*}[!htp]\centering
\scriptsize
\begin{tabular}{lrrrrrr}\toprule
English &Kayangan &Jenggala &Genggelang &Akar-Akar &Pemenang \\\midrule
Spear &tombak &cinokan &tar &tombak &t3r \\
There &ito &ito &ito &tino &Ito \\
Black &biron &biron &biron &pisak &biron \\
Worm &gumbor &lona &gumbor &gumbor &gumbor \\
Red &bonon &bonon &bonon &aban &bonon \\
Here &ite &ite &ite &tinI &Ite \\
Paddle &dayung &bose &dayung &bose &bose \\
Husband &sawa &sawa &sawa &sawa &kuronan \\
\bottomrule
\end{tabular}
\caption{Lexical variance of the Sasak language between the several villages in the district in Northern Lombok \citep{sarwadi2019lexical}}\label{tab:sasak }
\end{table*}

\subsubsection{Code-switching}
A code switcher or sometime called code-mixer is a person who uses two or more languages in conversation. Throughout a dialogue, a code-switcher shifts between two or more languages \citep{dougruoz2021surveycode}. This occurs frequently in Indonesian speech. People may merge their native languages with standard Indonesian in their speech, producing informal Indonesian. This casual version of Indonesian is spoken daily and is common on social networking websites. Even non-native speakers of the local languages can understand several regularly used code-mixed words, notably on social media. Code-switching can also develop in border zones when individuals are exposed to multiple languages, merging them together. Examples of code-switching is in table \ref{tab: code_switch } \citep{Aji2022}. 
\begin{table*}[!htp]\centering
\scriptsize
\begin{tabular}{lrrr}\toprule
Colloquial Indonesian (code-switching) &\multicolumn{2}{c}{Translation} \\\midrule
Quotenya (a,f) Bill Gates ini relevan banget (b) &\multicolumn{2}{c}{This quote from Bill Gates is very relevant} \\
Bilo(d) kito (d) pergi main lagi(f)? &\multicolumn{2}{c}{Shall we go play again?} \\
Ada yang ngetag(a,b) foto lawas(c) di FB &\multicolumn{2}{c}{Someone tagged an old photo on Facebook} \\
susah banget(b) &\multicolumn{2}{c}{very difficult.} \\
Ini teh(a) aksara jawa kenapa(f)? &\multicolumn{2}{c}{Why is this using Javanese script?} \\
\bottomrule
\end{tabular}
\caption{Examples of code switching in colloquial Indonesian taken from conversations on social media platforms.  alphabet code: (a) English, (b) Betawinese, (c) Javanese, (d) Minangkabau, (e) Sundanese,  and (f) Indonesian.\citep{Aji2022}}\label{tab: code_switch }
\end{table*}

\subsubsection{Unbalanced Local Indonesian monolingual data.}

Unlabeled corpora are essential for the construction of large language models, such as BERT \citep{bert} or GPT-2 by Open AI \citep{radford2019language-gpt2}. Indonesian-language unlabeled corpora such as Indo4B  and Indo4B-Plus \citep{indo4B-indonlu} mostly contain data in Indonesian, with the latter including a small amount of data in Javanese and Sundanese as well. For example, multilingual corpora such as CC–100 \citep{conneau2019unsupervised} contain only small amount (0.001\% of the total corpus size) of Javanese, while mC4 \citep{xue-etal-2021-mt5} contains only 0.6 million Javanese and 0.3 million Sundanese tokens out of a total of 6.3 million tokens. In addition, we compare the availability of data in Wikipedia to the number of speakers to determine data availability. When compared to European languages with equal numbers of speakers, there is significantly less information accessible for the languages spoken in Indonesia. According to Wikipedia, for example, Italian pages take up more than 3 GB of space whereas Javanese articles take up less than 50 MB, despite the fact that both languages have a similar number of native speakers. For the same reason, the articles in Sundanese are less than 25 megabytes in size, whereas languages with comparable numbers of speakers have more than 1.5 gigabytes in size. Similar tendencies can be observed in the majority of other Asian languages. In contrast to European languages with few speakers, the vast majority of other Indonesian native languages do not have Wikipedia pages. The majority of alternative sources for high-quality text data for other local languages of Indonesia (such as news websites) are written in Indonesian, making it extremely difficult to discover alternatives. Due to the extremely low number of speakers, long-tail languages have even fewer resources. Furthermore, the majority of languages in the long tail are predominantly spoken, making text data difficult to gather. These figures indicate how difficult it is to obtain unlabeled corpora for Indonesian local languages. This makes it hard to create robust pretrained language models for these languages, which have served as the basis for many modern NLP systems.

\subsection{Labeled Data For Downstream NLP Purposes}
Most work on Indonesian NLP has not publicly released the data or models, limiting reproducibility. Although recent Indonesian NLP benchmarks are addressing this issue, they mostly focus on the Indonesian language. Some of the widely used local languages, such as the top 10 regional languages by frequency of use (table \ref{tab:1}), have very few labeled data sets compared to Indonesian, while others have nearly none. The lack of such data sets makes it challenging to create NLP for local languages. However, constructing new labeled datasets is still challenging due to: (1) the lack of speakers of some languages; (2) the vast continuum of dialectical variation; and (3) the absence of writing standard in most local languages.

\subsection{Labeling Datasets with Twitter's Geotagging Service}
The data from Twitter is linguistically varied and contains tweets written in a large number of languages and dialects with limited resources. Having a Twitter geo tagging service that can help us collect large amounts of unlabeled text in low-resource languages. This text can then be used to enrich models for a variety of downstream natural language processing tasks \citep{twitterspoken}. This motivated the author to utilize Twitter databases to automatically crawling and label local Indonesian language data.

\subsection{NLP Transfer Learning}

In the field of Natural Language Processing (NLP), neural network topologies such as recurrent neural networks (RNNs) \citep{rnn} and convolutional neural networks (CNNs) \citep{cnn} have demonstrated significant improvements in performance. However, deep learning models in NLP perform poorly compared to deep learning models in Computer Vision. This poor development may be due to the lack of large tagged text datasets. Most labeled text datasets are too small to train deep neural networks, which have many parameters and overfitting occurs when trained on tiny datasets.

Deep learning in computer vision relies on transfer learning. Large labeled datasets, such as Imagenet, were used to train deep CNN-based models. Google's transformer model in 2017 repopularized natural language processing \citep{vaswani2017attention}. Transfer learning in NLP allows for high efficiency in a wide range of tasks. A deep learning model trained on a huge dataset can be used on another \citep{transfer}. This is a pre-trained deep learning model. Transformer-based models for NLP tasks arose quickly. Using transformer-based models has many benefits, the most important: Instead of processing input sequences token by token, these models can be expedited using GPUs. These models don't need labeled data. We simply need unlabeled text to train a transformer-based model. This model can classify, identify, and generate text. We will use BERT \citep{bert} to categorize text using a pre-trained BERT model.

BERT (Bidirectional Encoder Representations from Transformers) \citep{bert} is a large neural network design with 100 million to 300 million parameters. Overfitting occurs when a BERT model is trained on a tiny dataset. So, as a starting point, employ a pre-trained BERT model on a large dataset. The model can then be fine-tuned using our smaller dataset. Techniques for fine-tuning Train the complete architecture.
We can feed the output of the pre-trained model to a softmax layer. In this situation, the model's pre-trained weights are modified based on the new dataset. Frozen layer training – Partially training a pre-trained model is also possible. We can freeze the weights of the model's first layers while retraining only the upper levels. We can test how many layers to freeze and how many to train. Freeze the model's layers and add our own neural network layers to train a new model. During model training, just the associated layers' weights will be updated. 

\section{Proposed Framework}

As discussed in Chapter \ref{cap:challenge}, the challenges to establishing a dataset for regional languages are: (1) Resources are limited (2) Differences Between Dialects (3) code-switching (4) too many variations of writing and no writing standards. Therefore, we suggest that a framework is required to address the aforementioned issues. This work has multiple sides that can be completed; nevertheless, We will focus on developing a framework for efforts to automatically generate, collect, and classify datasets in tweets from 33 different provincial capitals retrieved over three years (2017 to 2020).

\begin{figure}[thp]
\includegraphics[width=\columnwidth]{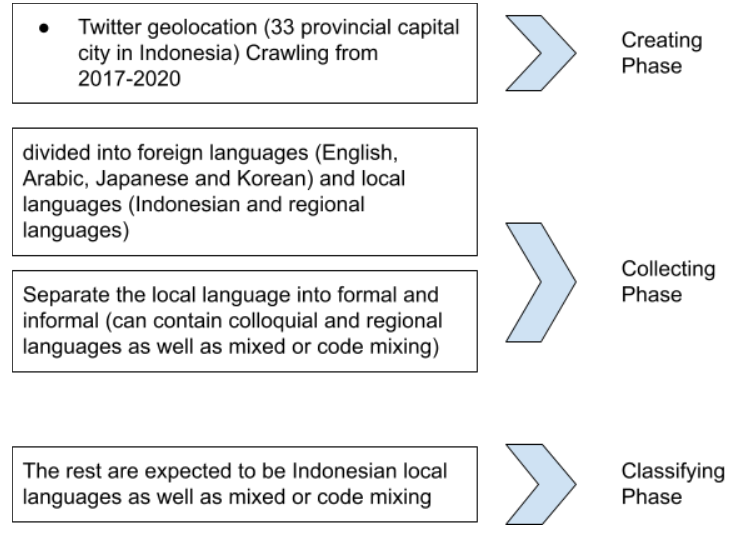}
\caption{the work will have three parts. First, create as many geotagged tweets as possible.Second, Filtration (Unsupervised) classified by foreign, local, and formal/colloquial languages. Third part Classification Phase (Supervised) detects tweet origin.}
\label{fig:2}
\end{figure}

As seen in Figure \ref{fig:2}, we will divide the task into three parts, the first of which is the creation phase: developing a monolingual local language from Twitter by collecting as many tweets as possible that are automatically labeled with geographic locations (33 cities crawled for three years). The second phase is the Filtration Phase (Unsupervised). During this phase, Twitter is filtered between foreign and local languages, and then the local language is filtered between formal and colloquial languages. The third phase, Classification Phase (Supervised), identifies the location from which a tweet is sent.


\subsection{Phase 1: Creating Monolingual dataset for model pretraining efforts}
Language evolves with its speakers. Larger-spoken languages will have a bigger digital presence, which will encourage more written forms of communication, whereas smaller-spoken languages will stress the spoken form. Some regional languages are also declining, with speakers preferring to adopt Indonesian over their own mother tongue. On the other hand, there are residents who live in isolation and speak the local language but they lack proficiency in the use of technology. Twitter users are extremely uncommon in this society.

Twitter is a medium that can be accessed and is quite popular in Indonesia, and it can also be used as an unsupervised automatic dataset annotation. The majority of tweet data in Indonesian regions use Indonesian, English, mixed languages, and informal language, while local language is used the least.

\begin{figure*}[ht]
\centering
     \includegraphics[width=2.0\columnwidth]{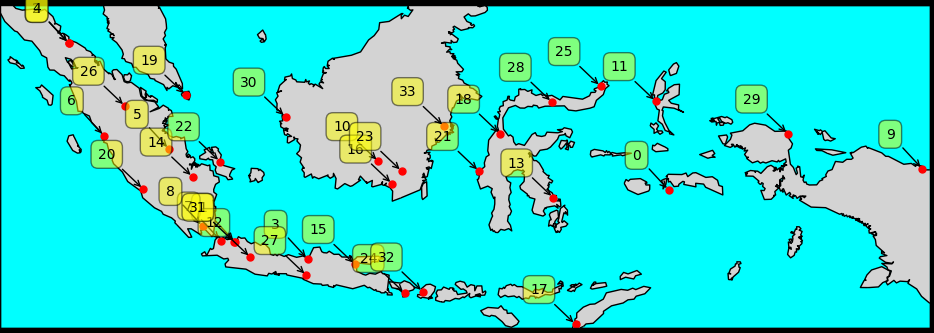}
      \caption{We Crawl Tweet from 33 Province Capital City, From 2017 to 2020}
       \label{fig:3}
\end{figure*}

The issue is that only a few tweets have geolocation data revealing where the tweet was sent. Previous research \citep{geotagging} indicates that around 0.85\% of tweets are geotagged, indicating that the longitude and latitude coordinates of the tweeter at the time the tweet was written are logged. Within three years of crawling Twitter, we obtained surprisingly little material. Figure \ref{fig:3} illustrates the geotagging map that we created.


\subsection{Phase 2: Collection}
In this second phase, we employ Fasttext to distinguish between foreign and local languages (including Indonesian). Not all foreign languages are utilized often in Indonesia, for instance. Although Mandarin is one of the most widely spoken languages in the world, it is infrequently used in Indonesia. Therefore, we only filter English, Japanese, Korean, and Arabic, which are the most popular foreign languages.

We employ filters to supervise the classification of formal Indonesian against non-formal languages (which can include colloquial, mixed, and regional languages) since it is easier to filter formal Indonesian. Therefore, we construct a training set consisting of formal and informal language.

Most language detectors utilize a method to identify the language being used; this becomes problematic if the language is a low-resource language, even if it is blended with another language. For instance, between 20 and 30 percent of modern Javanese speakers combine Javanese with Indonesian or other languages. Javanese is the second most commonly spoken language in Indonesia. The majority of Indonesians speak both Indonesian and their native tongue. Problematically, while communicating in writing, especially via social media, Indonesians tend to utilize colloquial Indonesian rather than their native tongue.

Instead of using language detection, we will use negative detection, that is, collect informal tweets, namely tweets that are not formal Indonesian, and not foreign languages.

\subsection{Phase 3: Classifying}

Many Indonesian regional languages are spoken and lack an orthographic system. Some indigenous languages have their own historic writing systems inherited from the Jawi or Kawi alphabet; standard transliterations into the Roman alphabet for some (e.g., Javanese and Sundanese) are not generally known and practiced \citep{indonesiakerancuan}. Because the sound is the same, several words have multiple romanized orthography that is mutually recognizable. Indonesian regional languages have diverse written forms. 

For NLP systems that use word-based representation, this variation increases vocabulary sizes and presents the issue of matching spelled representations of the same term. This phase is the most difficult one.

After collecting local, mixed, and informal language data, we will attempt to fine-tune the BERT model generated during pretraining. Here, we annotate supervised training data with the location where the tweet was made. Despite the fact that these tweets were collected in 33 locations over the course of three years, the resulting data is insufficient and unbalanced.

\section{Experiment Setup}

\begin{figure}[thp]
\includegraphics[width=\columnwidth]{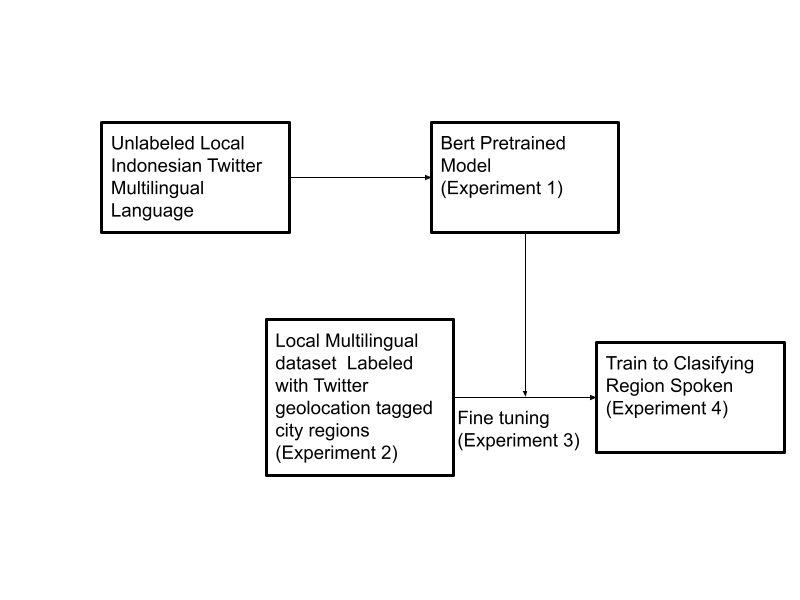}
\caption{Experimental Setup Chart}
\label{fig:4}
\end{figure}
Before doing the experiment, we collected, cleaned, and annotated Twitter data with the location where Twitter was created. The statistics table \ref{tab:statistics} displays data statistics, including the date and location of data crawling, as well as total data tabulation.

\begin{table}[!htp]\centering
\scriptsize
\begin{tabular}{lrr}\toprule
Data Statistics &Value \\\midrule
Crawl time period &2017-2020 \\
Cities (Table 8) &33 Cities \\
 Tweet total &1,326,099 \\
 foreign language &271,861 (20.5\%) \\
 formal Indonesian &131,843 (9.9\%) \\
 informal Indonesian &922,755 (69.6\%) \\
\bottomrule
\end{tabular}
\caption{Data statistics, including the date and location of data crawling, as well as total data tabulation.}\label{tab:statistics}
\end{table}

\subsection{Experiment 1}
We chose BERT as our pre-training dataset because it has been extensively investigated by Indonesian scholars. The first experiment will determine whether an Indonesian text is formal or colloquial. Using three pretrained BERT models, namely indolem-indobert-uncased \citep{koto2020indolem}, indonesia-bert-base-522M \citep{indo4B-indonlu}, and indobertweet-base-uncased \citep{koto2021indobertweet}.


\begin{table*}[ht!]
\centering
\scriptsize
\begin{tabular}{lrrrrrrrr}\toprule
\centering
&\multicolumn{2}{c}{Precission} &\multicolumn{2}{c}{Recall} &\multicolumn{2}{c}{F1-Score} &Acuracy \\\cmidrule{2-8}
Existing Pretrained Bert Datasets &Formal &Informal &Formal &Informal &Formal &Informal & \\\midrule
indolem indobert uncased &0.8 &0.85 &0.86 &0.78 &0.83 &0.81 &0.82 \\
indonesia-bert-base-522M &0.84 &0.86 &0.87 &0.84 &0.85 &0.85 &0.85 \\
indobertweet-base-uncased &\textbf{0.86} &\textbf{0.9} &\textbf{0.91} &\textbf{0.86} &\textbf{0.89} &\textbf{0.88} &\textbf{0.88} \\
\bottomrule
\end{tabular}
\caption{ Formal Indonesian and colloquial Indonesian were classified by comparing the outcomes of three previously trained BERT models}
\label{tab:indobert}    
\end{table*}

1922 of 3844 manually annotated tweets have formal language, while the remaining 1922 contain colloquial expressions. The data is then separated into 0.70 training data, 0.15 testing data, and 0.15 validation data.

By executing it for 50 fine-tuning epochs. With dropout layer = 0.1, activation function = ReLU, Layer 1 = Linear (768,521), Layer 2 = Linear (512,2), and the final layer's activation function is softmax, the text is classified as formal or informal Indonesian. The acquired findings are then presented in table \ref{tab:indobert}. 

From table \ref{tab:indobert}, it can be concluded that the indobertweet-base-uncased dataset is optimal for classifying formal and informal language in our dataset. This makes sense, as the dataset employs a mixed Indonesian Twitter dataset as the pretraining dataset.

\subsection{Experiment 2}
\label{sec:exp2}
After achieving the optimal pretrained and fine-tuned model in the first experiment, the second experiment will consist of acquiring a Twitter dataset incorporating local languages. This is achieved by eliminating tweets containing frequently-used foreign languages on Indonesian local Twitter. 

\begin{figure}[!htp]
\centering
   \includegraphics[width=0.9\linewidth]{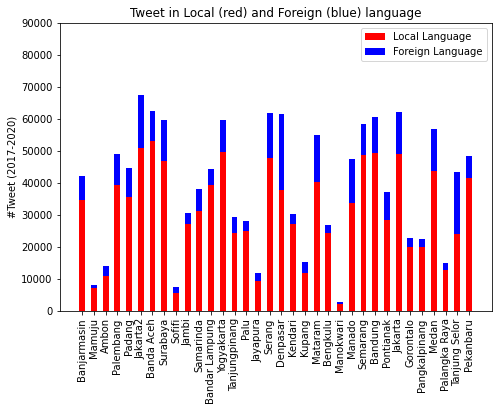}

\caption{Experiment 2: Filtering Foreign Language We filtered frequently used foreign languages on Indonesian Twitter, such as English, Arabic, Japanese, and Korean.}
\label{fig:foreign}
\end{figure}

As indicated in figure \ref{fig:3} and table, geotagged tweets are collected from 33 distinct regions in Indonesia. Facebook's FastText is used to reduce the number of tweets written in other languages that are commonly spoken in Indonesia, such as English, Japanese, Korean, and Arabic. Then, the data is depicted in graph in figure \ref{fig:foreign}.

The raw data includes formal Indonesian, informal Indonesian, and local languages in locations where Twitter supports geotagging. This was accomplished by filtering frequently used foreign languages on Twitter.

\subsection{Experiment 3}
\begin{figure}[!htp]
\centering
   \includegraphics[width=0.9\linewidth]{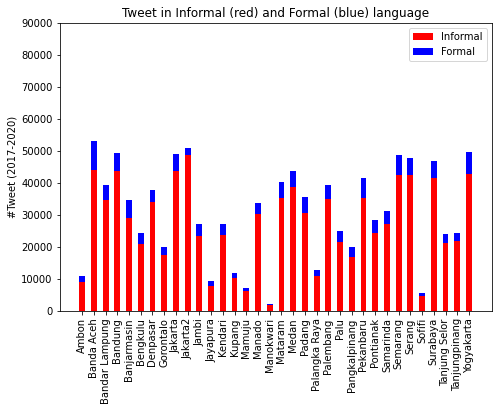}

\caption{Experiment 3: Filtering Formal Language, We filtered formal languages on Indonesian Twitter since they don't represent local languages and dialects. }
\label{fig:formal}
\end{figure}

In the first experiment, it was determined that the use of pretrained BERT indobertweet-base-uncased was the most applicable for Twitter dataset, and a model was developed to differentiate between formal and informal languages in experiment 2 in the section \ref{sec:exp2}. In this experiment, the model from the first experiment was utilized to differentiate the use of formal and informal language in distinct Indonesian regions. This experiment produced an intriguing pattern, as depicted in Figure \ref{fig:formal}. Formal Indonesian is a language that is mastered by all regions in Indonesia. However, in this low-resource experiment, formal language is regarded as noise because it does not represent a particular region where local languages and dialects are used. Therefore, formal language is separated from informal language used in the region.

The line between formal and informal language is sometimes quite thin, and one could argue that it is also subjective. This presents its own unique set of issues when attempting to differentiate between the two. As a consequence, the degree of accuracy is not quite as high. On the other hand, one may say that this model is highly effective when it comes to filtering.

\subsection{Experiment 4}
Given the mixed data of informal and local languages on Twitter discovered in Experiment 3, the final model is constructed to categorize the collection of tweets by the region from which they originated. This is used to determine which local languages are spoken in the area. This is possible due to Twitter's geolocation tagging feature, which may be used to annotate the tweet's location.

Using machine learning, we identify very few original statements in a given location; instead, there is code mixing or terms from many languages. It is not possible to extract terms that exclusively use local languages based on this research because Twitter users are often intended for the broader public, hence the majority use national and mixed languages (colloquial languages). To be able to entirely isolate the local language, more advanced approaches or even hand annotation are required; however, this is not an option due to its high cost.

\section{Analysis of Experimental Outcomes}
There are interesting findings from the separation of formal and informal (including regional languages), with manual searches found that there are only less than 1\% of tweets that use regional languages. This finding is quite surprising, because the authors expect more tweets in regional languages in certain regions. Even in areas with small tweet volumes, such as Manukwari, most tweets are dominated by Indonesian, both formal and informal. So that more aggressive filtering is needed to remove colloquial language with more sophisticated techniques.

Case study, the city of Surabaya, where the author currently lives, and the author's native language is Javanese, we conducted a manual search, using Twitter in Javanese in the city of Surabaya, which should be dominated by Javanese and using Javanese in daily conversation, but we found that less than 1 percent of tweets used the local language, and the rest were a mixture of languages. It indicates that \citet{anindyatri_mufidah_2020} claim that the use of regional languages is beginning to decline is very credible.

\section{Future Works}

Expensive GPU requirements limit our research. Instead of establishing another huge model, we advise developing lightweight and fast neural architectures, such as by distillation \citep{jiao2019tinybert}, model factorization, or model pruning \citep{voita2019analyzing}.  Non-neural approaches remain popular among Indonesian academics. Further research on the trade-off between model efficiency and quality is also interesting.

\section{Conclusion}
The framework that we built has the ability to reduce and filter out non-local languages by up to 30.4\%. This has never been done by researchers in Indonesia; yet, 30\% is still lacking because 70\% of these tweets are a mixture of colloquial language and code switching to dominate tweets. Less than one percent of the tweet is local language (we manually search with random samples).
There are many things that can be explored in the case of collecting datasets for local languages in Indonesia. We need even more aggressive filtering, because to distinguish whether an informal sentence is colloquial, code-mixing or pure regional language, more data is needed, we need to enter manual data by native speakers, this cannot be done by the author due to lack of resources.

\bibliography{anthology,custom}

\begin{thebibliography}{41}
\expandafter\ifx\csname natexlab\endcsname\relax\def\natexlab#1{#1}\fi

\bibitem[{Aji et~al.(2022)Aji, Winata, Koto, Cahyawijaya, Romadhony, Mahendra,
  Kurniawan, Moeljadi, Prasojo, Baldwin, Lau, and Ruder}]{Aji2022}
Alham~Fikri Aji, Genta~Indra Winata, Fajri Koto, Samuel Cahyawijaya, Ade
  Romadhony, Rahmad Mahendra, Kemal Kurniawan, David Moeljadi, Radityo~Eko
  Prasojo, Timothy Baldwin, Jey~Han Lau, and Sebastian Ruder. 2022.
\newblock \href {https://doi.org/10.48550/ARXIV.2203.13357} {One country, 700+
  languages: Nlp challenges for underrepresented languages and dialects in
  indonesia}.

\bibitem[{Anderbeck(2012)}]{anderbeck2012malayic}
Karl Anderbeck. 2012.
\newblock The malayic-speaking; orang laut dialects and directions for
  research.
\newblock \emph{Wacana}, 14(2):265--312.

\bibitem[{Anindyatri and Mufidah(2020)}]{anindyatri_mufidah_2020}
Anisya~Oktaviana Anindyatri and Imarotul Mufidah. 2020.
\newblock \emph{Gambaran kondisi vitalitas bahasa daerah di Indonesia:
  berdasarkan data Tahun 2018 – 2019.}
\newblock Pusat Data Informasi dan Teknologi, Kementerian Pendidikan dan
  Kebudayaan.

\bibitem[{Apriani et~al.(2016)Apriani, Sujaini, and
  Safriadi}]{apriani2016pengaruh}
Tri Apriani, Herry Sujaini, and Novi Safriadi. 2016.
\newblock Pengaruh kuantitas korpus terhadap akurasi mesin penerjemah statistik
  bahasa bugis wajo ke bahasa indonesia.
\newblock \emph{JUSTIN (Jurnal Sistem Dan Teknologi Informasi)}, 4(1):168--173.

\bibitem[{Banaei et~al.(2020)Banaei, Lebret, and Aberer}]{twitterspoken}
Mohammadreza Banaei, R{\'e}mi Lebret, and Karl Aberer. 2020.
\newblock Spoken dialect identification in twitter using a multi-filter
  architecture.
\newblock \emph{arXiv preprint arXiv:2006.03564}.

\bibitem[{Conneau et~al.(2019)Conneau, Khandelwal, Goyal, Chaudhary, Wenzek,
  Guzm{\'a}n, Grave, Ott, Zettlemoyer, and Stoyanov}]{conneau2019unsupervised}
Alexis Conneau, Kartikay Khandelwal, Naman Goyal, Vishrav Chaudhary, Guillaume
  Wenzek, Francisco Guzm{\'a}n, Edouard Grave, Myle Ott, Luke Zettlemoyer, and
  Veselin Stoyanov. 2019.
\newblock Unsupervised cross-lingual representation learning at scale.
\newblock \emph{arXiv preprint arXiv:1911.02116}.

\bibitem[{Devlin et~al.(2018{\natexlab{a}})Devlin, Chang, Lee, and
  Toutanova}]{https://doi.org/10.48550/arxiv.1810.04805}
Jacob Devlin, Ming-Wei Chang, Kenton Lee, and Kristina Toutanova.
  2018{\natexlab{a}}.
\newblock \href {https://doi.org/10.48550/ARXIV.1810.04805} {Bert: Pre-training
  of deep bidirectional transformers for language understanding}.

\bibitem[{Devlin et~al.(2018{\natexlab{b}})Devlin, Chang, Lee, and
  Toutanova}]{bert}
Jacob Devlin, Ming{-}Wei Chang, Kenton Lee, and Kristina Toutanova.
  2018{\natexlab{b}}.
\newblock \href {http://arxiv.org/abs/1810.04805} {{BERT:} pre-training of deep
  bidirectional transformers for language understanding}.
\newblock \emph{CoRR}, abs/1810.04805.

\bibitem[{Dewi et~al.(2020)Dewi, Santoso, Ubaidi, and
  Setyaningsih}]{dewi2020combination}
Nindian~Puspa Dewi, Joan Santoso, Ubaidi Ubaidi, and Eka~Rahayu Setyaningsih.
  2020.
\newblock Combination of genetic algorithm and brill tagger algorithm for part
  of speech tagging bahasa madura.
\newblock \emph{Proceeding of the Electrical Engineering Computer Science and
  Informatics}, 7(2):38--42.

\bibitem[{Do{\u{g}}ru{\"o}z et~al.(2021)Do{\u{g}}ru{\"o}z, Sitaram, Bullock,
  and Toribio}]{dougruoz2021surveycode}
A~Seza Do{\u{g}}ru{\"o}z, Sunayana Sitaram, Barbara~E Bullock, and
  Almeida~Jacqueline Toribio. 2021.
\newblock A survey of code-switching: Linguistic and social perspectives for
  language technologies.
\newblock In \emph{The Joint Conference of the 59th Annual Meeting of the
  Association for Computational Linguistics and the 11th International Joint
  Conference on Natural Language Processing (ACL-IJCNLP 2021)}. Association for
  Computational Linguistics.

\bibitem[{{Dukcapil Kemendagri}(2022)}]{dukcapil-kemendagri-2022}
{Dukcapil Kemendagri}. 2022.
\newblock \href
  {https://dukcapil.kemendagri.go.id/berita/baca/1032/273-juta-penduduk-indonesia-terupdate-versi-kemendagri#:~:text=Jakarta%20%2D%20Kemendagri%20melalui%20Direktorat%20Jenderal,Indonesia%20adalah%20273.879.750%20jiwa.}
  {{273 Juta Penduduk Indonesia Terupdate Versi Kemendagri}}.

\bibitem[{Fauzi and Puspitorini(2018)}]{fauzi2018dialect}
Andri~Imam Fauzi and Dwi Puspitorini. 2018.
\newblock Dialect and identity: A case study of javanese use in whatsapp and
  line.
\newblock In \emph{IOP Conference Series: Earth and Environmental Science},
  volume 175, page 012111. IOP Publishing.

\bibitem[{Hasbiansyah et~al.(2016)Hasbiansyah, Sujaini, and
  Safriadi}]{hasbiansyah2016tuning}
Muhammad Hasbiansyah, Herry Sujaini, and Novi Safriadi. 2016.
\newblock Tuning for quality untuk uji akurasi mesin penerjemah statistik (mps)
  bahasa indonesia-bahasa dayak kanayatn.
\newblock \emph{JUSTIN (Jurnal Sistem dan Teknologi Informasi)}, 4(1):209--213.

\bibitem[{Jiao et~al.(2019)Jiao, Yin, Shang, Jiang, Chen, Li, Wang, and
  Liu}]{jiao2019tinybert}
Xiaoqi Jiao, Yichun Yin, Lifeng Shang, Xin Jiang, Xiao Chen, Linlin Li, Fang
  Wang, and Qun Liu. 2019.
\newblock Tinybert: Distilling bert for natural language understanding.
\newblock \emph{arXiv preprint arXiv:1909.10351}.

\bibitem[{Joulin et~al.(2016)Joulin, Grave, Bojanowski, Douze, J{\'e}gou, and
  Mikolov}]{joulin2016fasttext}
Armand Joulin, Edouard Grave, Piotr Bojanowski, Matthijs Douze, H{\'e}rve
  J{\'e}gou, and Tomas Mikolov. 2016.
\newblock Fasttext.zip: Compressing text classification models.
\newblock \emph{arXiv preprint arXiv:1612.03651}.

\bibitem[{Kohler(2019)}]{kohler2019language}
Michelle Kohler. 2019.
\newblock Language education policy in indonesia: A struggle for unity in
  diversity.
\newblock \emph{The Routledge international handbook of language education
  policy in Asia}, pages 268--297.

\bibitem[{Koto and Koto(2020)}]{koto2020towards}
Fajri Koto and Ikhwan Koto. 2020.
\newblock Towards computational linguistics in minangkabau language: Studies on
  sentiment analysis and machine translation.
\newblock \emph{arXiv preprint arXiv:2009.09309}.

\bibitem[{Koto et~al.(2021)Koto, Lau, and Baldwin}]{koto2021indobertweet}
Fajri Koto, Jey~Han Lau, and Timothy Baldwin. 2021.
\newblock Indobertweet: A pretrained language model for indonesian twitter with
  effective domain-specific vocabulary initialization.
\newblock In \emph{Proceedings of the 2021 Conference on Empirical Methods in
  Natural Language Processing (EMNLP 2021)}.

\bibitem[{Koto et~al.(2020)Koto, Rahimi, Lau, and Baldwin}]{koto2020indolem}
Fajri Koto, Afshin Rahimi, Jey~Han Lau, and Timothy Baldwin. 2020.
\newblock Indolem and indobert: A benchmark dataset and pre-trained language
  model for indonesian nlp.
\newblock In \emph{Proceedings of the 28th COLING}.

\bibitem[{LeCun et~al.(1995)LeCun, Bengio et~al.}]{cnn}
Yann LeCun, Yoshua Bengio, et~al. 1995.
\newblock Convolutional networks for images, speech, and time series.
\newblock \emph{The handbook of brain theory and neural networks},
  3361(10):1995.

\bibitem[{Lewis(2022)}]{ethnologue}
M.~Paul Lewis, editor. 2022.
\newblock \emph{Ethnologue: Languages of the World}, 25 edition.
\newblock SIL International, Dallas, TX, USA.

\bibitem[{Liu et~al.(2020)Liu, Gu, Goyal, Li, Edunov, Ghazvininejad, Lewis, and
  Zettlemoyer}]{mbartliu}
Yinhan Liu, Jiatao Gu, Naman Goyal, Xian Li, Sergey Edunov, Marjan
  Ghazvininejad, Mike Lewis, and Luke Zettlemoyer. 2020.
\newblock \href {http://arxiv.org/abs/2001.08210} {Multilingual denoising
  pre-training for neural machine translation}.
\newblock \emph{CoRR}, abs/2001.08210.

\bibitem[{Maharani and Candra(2018)}]{maharani2018variasi}
Putu~Devi Maharani and Komang Dian~Puspita Candra. 2018.
\newblock Variasi leksikal bahasa bali dialek kuta selatan.
\newblock \emph{Mudra Jurnal Seni Budaya}, 33(1):76--84.

\bibitem[{Ningtyas et~al.(2018)Ningtyas, Sujaini, and
  Safriadi}]{ningtyas2018penggunaan}
Della~Widya Ningtyas, Herry Sujaini, and Novi Safriadi. 2018.
\newblock Penggunaan pivot language pada mesin penerjemah statistik bahasa
  inggris ke bahasa melayu sambas.
\newblock \emph{JEPIN (Jurnal Edukasi dan Penelitian Informatika)},
  4(2):173--178.

\bibitem[{Novitasari et~al.(2020)Novitasari, Tjandra, Sakti, and
  Nakamura}]{novitasari-etal-2020-cross}
Sashi Novitasari, Andros Tjandra, Sakriani Sakti, and Satoshi Nakamura. 2020.
\newblock \href {https://aclanthology.org/2020.sltu-1.18} {Cross-lingual
  machine speech chain for {J}avanese, {S}undanese, {B}alinese, and {B}ataks
  speech recognition and synthesis}.
\newblock In \emph{Proceedings of the 1st Joint Workshop on Spoken Language
  Technologies for Under-resourced languages (SLTU) and Collaboration and
  Computing for Under-Resourced Languages (CCURL)}, pages 131--138, Marseille,
  France. European Language Resources association.

\bibitem[{Pamolango(2012)}]{pamolango2012geografi}
Valantino~Ateng Pamolango. 2012.
\newblock Geografi dialek bahasa saluan.
\newblock \emph{PARAFRASE: Jurnal Kajian Kebahasaan \& Kesastraan}, 12(02).

\bibitem[{Pan et~al.(2017)Pan, Zhang, May, Nothman, Knight, and
  Ji}]{pan-etal-2017-cross}
Xiaoman Pan, Boliang Zhang, Jonathan May, Joel Nothman, Kevin Knight, and Heng
  Ji. 2017.
\newblock \href {https://doi.org/10.18653/v1/P17-1178} {Cross-lingual name
  tagging and linking for 282 languages}.
\newblock In \emph{Proceedings of the 55th Annual Meeting of the Association
  for Computational Linguistics (Volume 1: Long Papers)}, pages 1946--1958,
  Vancouver, Canada. Association for Computational Linguistics.

\bibitem[{Radford et~al.(2019)Radford, Wu, Child, Luan, Amodei, and
  Sutskever}]{radford2019language-gpt2}
Alec Radford, Jeff Wu, Rewon Child, David Luan, Dario Amodei, and Ilya
  Sutskever. 2019.
\newblock Language models are unsupervised multitask learners.

\bibitem[{Ruder et~al.(2019)Ruder, Peters, Swayamdipta, and Wolf}]{transfer}
Sebastian Ruder, Matthew~E. Peters, Swabha Swayamdipta, and Thomas Wolf. 2019.
\newblock \href {https://doi.org/10.18653/v1/N19-5004} {Transfer learning in
  natural language processing}.
\newblock In \emph{Proceedings of the 2019 Conference of the North {A}merican
  Chapter of the Association for Computational Linguistics: Tutorials}, pages
  15--18, Minneapolis, Minnesota. Association for Computational Linguistics.

\bibitem[{Rumelhart et~al.(1985)Rumelhart, Hinton, and Williams}]{rnn}
David~E Rumelhart, Geoffrey~E Hinton, and Ronald~J Williams. 1985.
\newblock Learning internal representations by error propagation.
\newblock Technical report, California Univ San Diego La Jolla Inst for
  Cognitive Science.

\bibitem[{Sarwadi et~al.(2019)Sarwadi, Mahsun, and
  Burhanuddin}]{sarwadi2019lexical}
Gita Sarwadi, Mahsun Mahsun, and Burhanuddin Burhanuddin. 2019.
\newblock Lexical variation of sasak kuto-kute dialect in north lombok
  district.
\newblock \emph{Jurnal Kata: Penelitian tentang Ilmu Bahasa dan Sastra},
  3(1):155--169.

\bibitem[{Sloan et~al.(2013)Sloan, Morgan, Housley, Williams, Edwards, Burnap,
  and Rana}]{geotagging}
Luke Sloan, Jeffrey Morgan, William Housley, Matthew Williams, Adam Edwards,
  Pete Burnap, and Omer Rana. 2013.
\newblock Knowing the tweeters: Deriving sociologically relevant demographics
  from twitter.
\newblock \emph{Sociological research online}, 18(3):74--84.

\bibitem[{Soeparno(2015)}]{indonesiakerancuan}
Soeparno. 2015.
\newblock Kerancuan fono-ortografis dan orto-fonologis bahasa indonesia ragam
  lisan dan tulis.

\bibitem[{Subali and Fatichah(2019)}]{subali2019kombinasi}
Made Agus~Putra Subali and Chastine Fatichah. 2019.
\newblock Kombinasi metode rule-based dan n-gram stemming untuk mengenali
  stemmer bahasa bali.
\newblock \emph{Jurnal Teknologi Informasi dan Ilmu Komputer}, 6(2):219--228.

\bibitem[{Suryani et~al.(2018)Suryani, Widyantoro, Purwarianti, and
  Sudaryat}]{suryani_sunda}
Arie~Ardiyanti Suryani, Dwi~Hendratmo Widyantoro, Ayu Purwarianti, and Yayat
  Sudaryat. 2018.
\newblock \href {https://doi.org/10.1145/3195634} {The rule-based sundanese
  stemmer}.
\newblock \emph{ACM Trans. Asian Low-Resour. Lang. Inf. Process.}, 17(4).

\bibitem[{Vander~Klok(2015)}]{vander2015dichotomy}
Jozina Vander~Klok. 2015.
\newblock The dichotomy of auxiliaries in javanese: Evidence from two dialects.
\newblock \emph{Australian Journal of Linguistics}, 35(2):142--167.

\bibitem[{Vaswani et~al.(2017)Vaswani, Shazeer, Parmar, Uszkoreit, Jones,
  Gomez, Kaiser, and Polosukhin}]{vaswani2017attention}
Ashish Vaswani, Noam Shazeer, Niki Parmar, Jakob Uszkoreit, Llion Jones,
  Aidan~N Gomez, {\L}ukasz Kaiser, and Illia Polosukhin. 2017.
\newblock Attention is all you need.
\newblock \emph{Advances in neural information processing systems}, 30.

\bibitem[{Voita et~al.(2019)Voita, Talbot, Moiseev, Sennrich, and
  Titov}]{voita2019analyzing}
Elena Voita, David Talbot, Fedor Moiseev, Rico Sennrich, and Ivan Titov. 2019.
\newblock Analyzing multi-head self-attention: Specialized heads do the heavy
  lifting, the rest can be pruned.
\newblock \emph{arXiv preprint arXiv:1905.09418}.

\bibitem[{Wilie et~al.(2020)Wilie, Vincentio, Winata, Cahyawijaya, Li, Lim,
  Soleman, Mahendra, Fung, Bahar, and Purwarianti}]{indo4B-indonlu}
Bryan Wilie, Karissa Vincentio, Genta~Indra Winata, Samuel Cahyawijaya,
  Xiaohong Li, Zhi~Yuan Lim, Sidik Soleman, Rahmad Mahendra, Pascale Fung,
  Syafri Bahar, and Ayu Purwarianti. 2020.
\newblock \href {https://aclanthology.org/2020.aacl-main.85} {{I}ndo{NLU}:
  Benchmark and resources for evaluating {I}ndonesian natural language
  understanding}.
\newblock In \emph{Proceedings of the 1st Conference of the Asia-Pacific
  Chapter of the Association for Computational Linguistics and the 10th
  International Joint Conference on Natural Language Processing}, pages
  843--857, Suzhou, China. Association for Computational Linguistics.

\bibitem[{Xue et~al.(2020)Xue, Constant, Roberts, Kale, Al{-}Rfou, Siddhant,
  Barua, and Raffel}]{mT5Xue2020}
Linting Xue, Noah Constant, Adam Roberts, Mihir Kale, Rami Al{-}Rfou, Aditya
  Siddhant, Aditya Barua, and Colin Raffel. 2020.
\newblock \href {http://arxiv.org/abs/2010.11934} {mt5: {A} massively
  multilingual pre-trained text-to-text transformer}.
\newblock \emph{CoRR}, abs/2010.11934.

\bibitem[{Xue et~al.(2021)Xue, Constant, Roberts, Kale, Al-Rfou, Siddhant,
  Barua, and Raffel}]{xue-etal-2021-mt5}
Linting Xue, Noah Constant, Adam Roberts, Mihir Kale, Rami Al-Rfou, Aditya
  Siddhant, Aditya Barua, and Colin Raffel. 2021.
\newblock \href {https://doi.org/10.18653/v1/2021.naacl-main.41} {m{T}5: A
  massively multilingual pre-trained text-to-text transformer}.
\newblock In \emph{Proceedings of the 2021 Conference of the North American
  Chapter of the Association for Computational Linguistics: Human Language
  Technologies}, pages 483--498, Online. Association for Computational
  Linguistics.

\end{thebibliography}
\bibliographystyle{acl_natbib}

\appendix

\section{Appendix}
\label{sec:appendix}

\begin{table*}[!htp]\centering
\scriptsize
\begin{tabular}{lrrrrrrrr}\toprule
& & & & & &\multicolumn{2}{c}{Indonesian} \\\cmidrule{7-8}
City &Latitute &Longitute & Raw Tweet & Foreign &Indonesian &Formal &Coloquial+Local \\\midrule
Ambon &-3.62553 &128.190643 &13998 &2930 &11068 &1859 &9209 \\
Banda Aceh &3.5912 &98.69175 &62618 &9477 &53141 &9208 &43933 \\
Bandar Lampung &-5.39714 &105.266792 &44377 &5046 &39331 &4752 &34579 \\
Bandung &-6.917464 &107.619125 &60724 &11459 &49265 &5522 &43743 \\
Banjarmasin &-3.318607 &114.594376 &42195 &7501 &34694 &5699 &28995 \\
Bengkulu &-3.577847 &102.34639 &26947 &2523 &24424 &3377 &21047 \\
Denpasar &-8.670458 &115.212631 &61463 &23770 &37693 &3737 &33956 \\
Gorontalo &0.699937 &122.446724 &22750 &2751 &19999 &2453 &17546 \\
Jakarta &-6.17511 &106.865036 &129594 &29647 &99947 &7319 &92628 \\
Jambi &-1.610123 &103.613121 &30747 &3668 &27079 &3634 &23445 \\
Jayapura &-2.591603 &140.669006 &11961 &2482 &9479 &1505 &7974 \\
Kendari &-3.99846 &122.512978 &30190 &3031 &27159 &3443 &23716 \\
Kupang &-10.1772 &123.607033 &15422 &3671 &11751 &1539 &10212 \\
Mamuju &-2.69919 &118.862106 &8041 &746 &7295 &886 &6409 \\
Manado &1.47483 &124.842079 &47531 &13845 &33686 &3386 &30300 \\
Manokwari &-0.861453 &134.062042 &2677 &345 &2332 &422 &1910 \\
Mataram &-8.597081 &116.100487 &55077 &14887 &40190 &4980 &35210 \\
Medan &3.595196 &98.672226 &56768 &12911 &43857 &5169 &38688 \\
Padang &-0.947083 &100.417183 &44670 &8881 &35789 &5171 &30618 \\
Palangka Raya &-2.216105 &113.913979 &15109 &2338 &12771 &1791 &10980 \\
Palembang &-2.976074 &104.775429 &49133 &9682 &39451 &4580 &34871 \\
Palu &-0.86791 &119.904655 &28067 &2926 &25141 &3571 &21570 \\
Pangkalpinang &-2.22487 &106.124649 &22534 &2437 &20097 &3358 &16739 \\
Pekanbaru &0.507068 &101.447777 &48397 &6845 &41552 &6148 &35404 \\
Pontianak &-0.02633 &109.342506 &37239 &8732 &28507 &4020 &24487 \\
Samarinda &-0.494823 &117.143616 &37991 &6717 &31274 &3915 &27359 \\
Semarang &-7.005145 &110.438126 &58337 &9424 &48913 &6282 &42631 \\
Serang &-6.110366 &106.163979 &61871 &14117 &47754 &5359 &42395 \\
Sofifi &0.734965 &127.561447 &7432 &1740 &5692 &978 &4714 \\
Surabaya &-7.257472 &112.75209 &59690 &12828 &46862 &5230 &41632 \\
Tanjung Selor &-2.69844 &115.057564 &43409 &19247 &24162 &2945 &21217 \\
Tanjungpinang &1.04912 &104.440659 &29399 &5140 &24259 &2507 &21752 \\
Yogyakarta &-7.79558 &110.369492 &59741 &10117 &49624 &6738 &42886 \\
\bottomrule
\end{tabular}
\caption{The tabulation of all experimental (2,3,4) results in a single data table}\label{tab:tabulate_all }
\end{table*}

\end{document}